\documentclass[letterpaper,10pt,journal,twoside]{IEEEtran}
\ifdefined\pdfsuppressptexinfo\pdfsuppressptexinfo=-1\fi
\ifdefined\pdfinfoomitdate\pdfinfoomitdate=1\fi
\ifdefined\pdftrailerid\pdftrailerid{}\fi
\ifdefined{/Creator () /Producer () /Trapped /Unknown}\fi
\usepackage{amsmath,amssymb,amsfonts}
\usepackage{graphicx}
\graphicspath{{figures/}}
\usepackage{textcomp}
\usepackage{xcolor}
\usepackage{booktabs}
\usepackage{url}
\urldef{\projecturl}\url{https://eddytac.github.io}
\newcommand{\sminus}{\scalebox{0.6}[1]{$\scriptstyle-$}}
\newsavebox{\tabnotebox}
\newcommand{\tabnote}[1]{\par\vspace{3pt}%
  \parbox{\wd\tabnotebox}{\footnotesize\raggedright #1}}
\usepackage{placeins}   %

\begin{document}
\title{EddyTac: A 6-Axis Inductive Tactile Sensor Using Coplanar PCB Coils and a Passive Conductive Plate}
\author{Jinho So$^{*}$, Ankush K. Dhawan$^{*}$, Max Burns, and Monroe Kennedy III%
\thanks{$^{*}$J. So and A. K. Dhawan contributed equally to this work.}%
\thanks{The authors are with the Department of Mechanical Engineering,
Stanford University, Stanford, CA 94305 USA. J. So is supported by a
Samsung PhD Fellowship.}%
\thanks{Project page: \projecturl}}

\markboth{}{So \MakeLowercase{\textit{et al.}}: EddyTac: A 6-Axis Inductive Tactile Sensor Using Coplanar PCB Coils and a Passive Conductive Plate}

\maketitle

\begin{abstract}
We present EddyTac, an inductive six-axis tactile sensor
that is low-cost, lightweight, and only 3.1~mm thick. Seven coplanar coils
on a standard printed circuit board sit beneath a three-blade conductive
plate embedded in silicone. The cast silicone stack is bonded to the board
and carries no wiring, and all molds and alignment fixtures are 3D printed.
Contact loads displace the plate over the coils in six degrees of freedom.
Using simulated coil inductances, we place and orient the plate edges so that
the plate's three translations and three rotations produce linearly
independent changes in the measured coil inductances near the rest pose.
A learned model maps the seven coil inductance changes to the contact wrench.
In a separate test session, the model had average root-mean-square errors of 0.15~N across force components and 1.6~mN$\cdot$m across moment components.
We demonstrate EddyTac in two applications, using two form factors that
scale the same design. An EddyTac sensor can trace handwriting using the
center of pressure, and a pair of fingertip sensors can localize a pinch grasp
along a bar within 4.3~mm of the intended position on average.\\
Project page: \projecturl
\end{abstract}


\section{Introduction}
\label{sec:introduction}

\begin{table*}[!tb]
\caption{
Comparison of six-axis tactile and force--torque sensors.
}
\label{tab:comparison}
\centering
\footnotesize
\setlength{\tabcolsep}{4pt}
\sbox{\tabnotebox}{%
\begin{tabular}{@{}llllcllll@{}}
\toprule
\textbf{Sensor} & \textbf{Transduction} & \textbf{Size (mm)} &
 \textbf{Mass (g)} & & \textbf{Load range} &
 \multicolumn{2}{l}{\textbf{Error}} & \textbf{Cost (\$)} \\
\midrule
ATI Nano17 \cite{ati_nano17} & Strain gauge & $\varnothing 17\times 14.5$ & 9.1 &
 $F$ & $(\pm 25,\, \pm 25,\, \pm 35)$ & NR &  & $\approx$5,000 \\
 & & & & $\tau$ & $(\pm 250,\, \pm 250,\, \pm 250)$ & &  & \\
\addlinespace[2pt]
Kim and Kim \cite{kim2025inductive} & Inductive & $\varnothing 85\times 36$ & 280 &
 $F$ & $(\pm 890,\, \pm 890,\, \pm 1435)$ & \%FS$_{\max}$ & $0.64$ & NR \\
 & & & & $\tau$ & $(\pm 27,\, \pm 27,\, \pm 45)\times 10^{3}$ & & $0.75$ & \\
\addlinespace[2pt]
OriCube \cite{xu2026oricube} & Inductive & $14.5\times 14.5\times 12.5$ & 4 &
 $F$ & $(\pm 23,\, \pm 23,\, 0\text{--}23)$ & Mean (\%FS) & $<1$ & NR$^*$ \\
 & & & & $\tau$ & $(\pm 125,\, \pm 125,\, \pm 160)$ & & $<1$ & \\
\addlinespace[2pt]
Kim \emph{et al.} \cite{kim2020fingertip} & Capacitive & $17\times 14\times 6.5$ & 2.6 &
 $F$ & $\pm 30$ each & Mean \%FSO & $(0.92,\, 0.83,\, 1.57)$ & NR$^*$ \\
 & & & & $\tau$ & $\pm 300$ each & & $(1.62,\, 1.71,\, 0.57)$ & \\
\addlinespace[2pt]
HexaTouch \cite{hexatouch} & Capacitive & $15\times 15\times 8$ & NR &
 $F$ & $(\pm 5,\, \pm 5,\, 0\text{--}25)$ & RMSE & $(0.14,\, 0.13,\, 0.13)$ & NR \\
 & & & & $\tau$ & $(\pm 100,\, \pm 100,\, \pm 50)$ & & $(2.48,\, 2.50,\, 0.95)$ & \\
\addlinespace[2pt]
CoinFT \cite{coinft} & Capacitive & $\varnothing 20\times 2$ & 2.0 &
 $F$ & $(\pm 5,\, \pm 5,\, 0\text{--}15)$ & RMSE & $(0.186,\, 0.134,\, 0.186)$ & 11 \\
 & & & & $\tau$ & $(\pm 100,\, \pm 100,\, \pm 80)$ & & $(1.114,\, 1.322,\, 0.810)$ & \\
\addlinespace[2pt]
Mao \emph{et al.} \cite{mao2025} & Thermal & $12\times 15\times 5$ & 0.30 &
 $F$ & $(\pm 5,\, \pm 5,\, 0\text{--}15)$ & RMSE & $(0.18,\, 0.15,\, 0.18)$ & NR$^*$ \\
 & & & & $\tau$ & $(\pm 60,\, \pm 60,\, \pm 20)$ & & $(0.58,\, 0.62,\, 0.26)$ & \\
\addlinespace[2pt]
LiVec finger \cite{livec} & Optical & $26.4\times 38.2\times 12$ & NR &
 $F$ & $(\pm 2,\, \pm 2,\, 0\text{--}12)$ & SD & $(0.112,\, 0.092,\, 0.139)$ & NR$^*$ \\
 & & & & $\tau$ & NR & & $(1.90,\, 1.54,\, 1.26)$ & \\
\addlinespace[2pt]
El-Azizi \emph{et al.} \cite{elazizi} & Optical & $\varnothing 27\times 20.8$ & 16.4 &
 $F$ & 0--5 each & MAE & $(0.06,\, 0.05,\, 0.07)$ & 100 \\
 & & & & $\tau$ & 0--200 each & & $(2.62,\, 2.68,\, 0.64)$ & \\
\addlinespace[2pt]
F-TOUCH \cite{ftouch} & Visuotactile & NR & NR &
 $F$ & $(\pm 1,\, \pm 1,\, 0\text{--}8)$ & RMSE & $(0.018,\, 0.016,\, 0.094)$ & NR \\
 & & & & $\tau$ & $(\pm 20,\, \pm 20,\, \pm 20)$ & & $(0.657,\, 1.566,\, 0.407)$ & \\
\midrule
\textbf{EddyTac (ours)} & \textbf{Inductive} & $\boldsymbol{\varnothing 25.8\times 3.1}$ & 3.2 &
 $F$ & $(9.3,\, 8.7,\, 34.5)$ & RMSE & $(0.110,\, 0.089,\, 0.265)$ & 9.21 \\
 & & & & $\tau$ & $(208,\, 187,\, 87)$ & & $(1.70,\, 2.09,\, 1.02)$ & \\
\bottomrule
\end{tabular}}
\usebox{\tabnotebox}
\tabnote{Forces are in N and moments in mN$\cdot$m, with components ordered
$(x, y, z)$. Sizes are length $\times$ width $\times$ thickness, or
$\varnothing$diameter $\times$ thickness for a circular sensor. RMSE is root-mean-square
error, MAE mean absolute error, SD standard deviation, and \%FS$_{\max}$ the
maximum error as a percentage of full scale. Kim \emph{et al.} normalize errors by the
reported sensor range. For EddyTac, the loads are the peak magnitudes on
each axis during training, and the errors are from real-time estimates in a
separate test session (Table~\ref{tab:ft_accuracy}). Test protocols vary across studies.
NR means not reported, and NR$^*$ that the paper calls the sensor low-cost
but gives no price. The ATI Nano17 price is for the sensor without its
data acquisition hardware~\cite{hendrich2020}. EddyTac's cost is its BOM cost per sensor at
1,000 units. El-Azizi \emph{et al.} give
the range that most of their loads fell in on each axis. The CoinFT and EddyTac sizes
are those of the circular sensing area. Their readout electronics occupy an
additional $10\times 11$~mm tab and a $25.8\times 10.8$~mm strip of the
same board.}
\end{table*}

Contact-rich tasks such as in-hand manipulation, tool use, and insertion
depend not only on how hard a robot presses on an object,
but on where and how it pushes, twists, and slides against it. External cameras provide geometric context,
but the contacting bodies can occlude the contact area. Tactile sensors measure these interactions at the contact surface directly.
They can improve robotic grasp stability and slip detection \cite{romano2011},
contact-rich assembly \cite{dong2021insertion}, and the success rate of learned
manipulation policies, particularly when visual observations are
degraded \cite{touchwild,tactilevla,manifeel,osmo}.

The simplest tactile sensors measure only the normal force $F_z$, which
determines how hard the object presses on the sensor. The shear forces
$F_x$ and $F_y$ give the direction of the tangential load. Their magnitude
relative to the normal force helps assess slip when the contact friction is
known \cite{goyal1991}. On a known contact plane, the measured forces and
tilting moments $\tau_x$ and $\tau_y$ locate the center of pressure (CoP)
\cite{bicchi1993}. The moment $\tau_z$ gives the twist about the contact normal when an object is pinched off its center of mass. These six components together form the contact wrench
\cite{murray1994}.

Many tactile skins measure only the normal force \cite{touchwild}, and those
that add shear forces still miss the contact moments
\cite{tactilevla,opentac,uskin,reskin}. Six-axis sensors
often use flexures, upright structures, or cameras that add thickness at the
contact and take up space inside a hand or gripper. Thin planar coil sensors,
in turn, resolve at most three load components at a time
\cite{wang2018triaxis,zhang2024rmfs}. In contrast, a thin six-axis sensor can fit
both a human data-collection device and a robot finger, giving the two
systems the same tactile measurement, which narrows the embodiment gap in
human-to-robot skill transfer \cite{osmo,huang2026flexitac}. Table~\ref{tab:comparison} compares the
size, load range, and accuracy of existing six-axis sensors.

\begin{figure}[!t]
\centering
\includegraphics[width=\columnwidth]{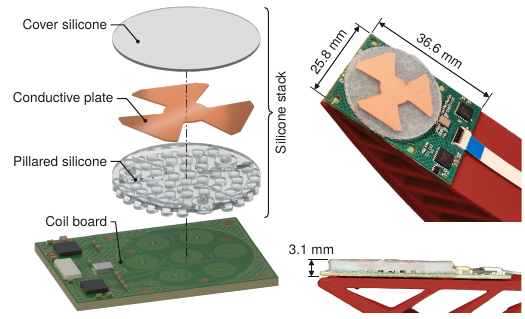}
\caption{
\textbf{Exploded view of EddyTac.} Left describes the sensor stackup of the cover silicone, the conductive plate, the pillared silicone,
and the coil board. Right displays the assembled sensor on a gripper finger in angled and side views.
The sensor is compact and thin with dimensions of $25.8\times36.6\times 3.1$~mm.}
\label{fig:teaser}
\end{figure}

We present EddyTac (Fig.~\ref{fig:teaser}), a thin inductive tactile sensor that addresses
these limitations using only planar parts: a coil board, a
conductive plate, and a silicone
stack that holds the conductive plate above the coils.
EddyTac orients six edges over coplanar
coils so that the plate's three translations and
three rotations produce linearly independent inductance responses near the
rest pose.
The conductive plate needs no wiring, and the readout
electronics sit outside the sensing area rather than beneath it. We
verify the design in simulation and
calibrate the sensor against a reference force/torque (F/T) sensor, achieving
root-mean-square errors (RMSEs) of 0.09--0.26~N for forces and 1.0--2.1~mN$\cdot$m for moments.

We make three contributions.
\begin{itemize}
\item A conductive plate geometry whose edges over six coplanar coils make all
six plate motions distinguishable, with the edge positions and angles
optimized for the weakest direction (Sec.~\ref{sec:principle}).
\item A 3.1~mm thick sensor made of a standard PCB and a cast silicone stack,
with all molds and alignment fixtures 3D printed (Sec.~\ref{sec:build}).
\item Six-axis calibration evaluated on a separate test session, and
demonstrations that locate the CoP and, with fingertip
sensors on the thumb and index finger, the grasp position along a bar
(Secs.~\ref{sec:device} and~\ref{sec:applications}).
\end{itemize}

We plan to open-source the sensor design to support replication and
adaptation by the broader robotics and sensors communities.

\section{Related Work}
\label{sec:related}

\subsection{Tactile Sensors for Manipulation}
\label{sec:related:tactile}

Tactile sensors for manipulation use many sensing modalities, and
their raw signals, such as resistance, magnetic field, or images,
differ from sensor to sensor. The forces and moments they estimate, by
contrast, are the same physical quantities for every sensor, so we
compare them here by the wrench components they measure. Some report only normal pressure, such as the
dense array of \cite{touchwild}. Others add shear force, for up to
three force components, including the gripper sensor of
\cite{tactilevla}, OpenTac \cite{opentac}, uSkin \cite{uskin}, the
sensor of Wang \emph{et al.} \cite{wang2018triaxis}, and ReSkin \cite{reskin}. AnySkin reports similar
signals but does not calibrate them to force \cite{anyskin}. These sensors report
at most three force components and do not measure contact moments.

\subsection{Six-Axis Force--Torque Sensing}
\label{sec:related:sixaxis}

Unlike the tactile sensors above, six-axis F/T sensors measure
all six components of the contact wrench \cite{cao2021review}.
To separate all six components, the readings must respond to each
component differently, so the sensing elements are often oriented along
several directions, usually on a three-dimensional structure.
Commercial industrial sensors such as the ATI Nano17 place strain gauges on a
metal flexure inside a housing \cite{ati_nano17}. Capacitive designs
measure the capacitance between electrodes in a standalone unit
\cite{kim2017novel} or around an eccentric cylinder at fingertip scale
\cite{kim2020fingertip}. Inductive designs place coils horizontally and
vertically around a T-beam \cite{kim2025inductive} or fold a flexible
coil board around a cube, as in OriCube \cite{xu2026oricube}.
DenseTact~2.0 instead images the deformation of a fingertip with an
internal camera \cite{densetact2}.
Commercial units bond miniature
strain gauges to micro-machined metal flexures and need dedicated
signal-conditioning electronics, so they typically cost several
thousand U.S. dollars \cite{veliky2024}.

\subsection{Thin Six-Axis Sensors}
\label{sec:related:thin}

The thin fingertip sensors in Table~\ref{tab:comparison} use several
fabrication methods. CoinFT uses a spin-coated silicone layer and film
spacers to bond its two sensing boards \cite{coinft}. HexaTouch uses slots
in a rigid support to align its molded elastomer to within 0.1~mm
\cite{hexatouch}. Mao \emph{et al.} use photolithography and sputtering
to fabricate thin-film thermistors \cite{mao2025}. These steps require a
spin coater, cleanroom processes such as photolithography and sputtering, or
alignment to 0.1~mm.

\subsection{Inductive and Eddy-Current Tactile Sensors}
\label{sec:related:inductive}

Eddy-current sensing is a standard way to measure displacement without
contact \cite{fleming2013,sophian2017,ti_target}. The alternating
magnetic field of a coil induces currents in a nearby conductive target, and changes in the gap
and lateral overlap between them change the coil's inductance
\cite{wang2018triaxis}. The target itself is passive and needs no
wiring, so all electrical connections stay on a fixed coil board, which
simplifies fabrication and assembly.
Previous inductive tactile sensors have used metal,
magnetorheological-elastomer, and liquid-metal targets
\cite{wang2018triaxis,yeh2019cmos,kawasetsu2018mre,kawasetsu2019liquid},
as well as coupled coil layers \cite{chen2025dual} and compact
multi-axis layouts \cite{zhang2024rmfs,xu2024sdmfs}. However, each resolves at most three load components in a
given configuration. The sensor of Zhang \emph{et al.} can be
reconfigured to measure $F_z$, $\tau_{x}$, and $\tau_{y}$ instead of the three
forces, but not all six at once \cite{zhang2024rmfs}. Six-axis inductive
sensors have used upright coils \cite{kim2025inductive} and folded coil
arrays \cite{xu2026oricube}.

\subsection{Summary}
\label{sec:related:summary}

EddyTac differs from these sensors in four ways. Unlike the tactile skins
of Sec.~\ref{sec:related:tactile}, it measures all six components of the
contact wrench. Unlike the six-axis sensors of Sec.~\ref{sec:related:sixaxis},
it needs no three-dimensional structure and uses only planar parts. The silicone stack
is cast in 3D-printed molds and placed on the coil board with a 3D-printed
aligner, without spin coating or cleanroom processes, and calibration
absorbs small placement errors (Sec.~\ref{sec:build:fab}). Finally, where
earlier six-axis inductive designs use upright coils or folded coil arrays,
EddyTac keeps every coil in one plane and shapes the edges of a
passive conductive plate so that the coils distinguish all six plate
motions (Sec.~\ref{sec:principle}).

\FloatBarrier
\section{Principle and Conductive Plate Design}
\label{sec:principle}
\label{sec:design}
An applied wrench displaces the sensor's conductive plate as a rigid
body in up to six degrees of freedom (DOF). Changes in coil inductance monitor this
displacement, and a learned model then converts these measured changes into a wrench estimate. This section describes the sensing principle and our specific design.  

\begin{figure*}[!tb]
\centerline{\includegraphics[width=\textwidth]{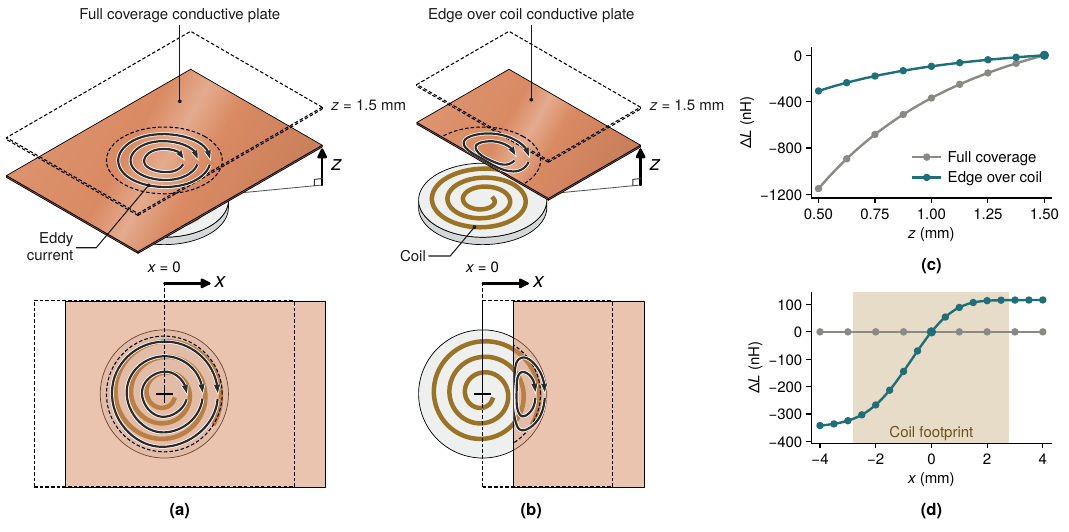}}
\caption{\textbf{One coil under full plate coverage and under a plate edge.} Calculations are made using the thin-sheet PEC model at a $z=1.5$~mm rest gap. (a) Full coverage. (b) An edge at the coil center ($x=0$). The isometric views press the conductive plate to $z=0.75$~mm, the top views slide it $1.5$~mm along $x$, and dashed outlines mark the rest location. Dark lines show the eddy current paths, which follow the edge when there is one. (c) Change in effective inductance $\Delta L$ from each plate's rest pose ($x=0$, $z=1.5$~mm) as the gap varies. The edge lowers the sensitivity at rest from $513$ to $130$~nH/mm. (d) $\Delta L$ against the lateral position $x$, $126$~nH/mm with the edge and zero under full coverage.}
\label{fig:mech}
\end{figure*}

\subsection{Single-Coil Eddy-Current Sensing}
\label{sec:principle:single}

A coil driven by an alternating current generates an alternating magnetic
field. By Faraday's law, this induces eddy currents in the nearby conductive
plate. By Lenz's law, these currents oppose the change in magnetic flux
\cite{smythe1968}. Their field reduces the coil's effective inductance,
as in the analytical solution for a circular coil above a conducting slab
\cite{dodd1968}. The conductive plate's eddy currents act as a closed
loop with inductance $L_{\mathrm{plate}}$, coupled to the coil across
the air gap by a mutual inductance $M$. When the loop's reactance dominates
its resistance ($\omega L_{\mathrm{plate}} \gg R_{\mathrm{plate}}$, with
$\omega$ the angular drive frequency), the coil's
effective inductance falls from its value $L_{\mathrm{coil}}$ without the plate to
\begin{equation}
    L_{\mathrm{eff}} = L_{\mathrm{coil}} -
    \frac{M(\mathbf{q})^2}{L_{\mathrm{plate}}}.
    \label{eq:leff}
\end{equation}
The mutual inductance $M(\mathbf{q})$ is the pose-dependent term, where $\mathbf{q}$ is the full six-DOF pose of the conductive plate
relative to the coil. However, a single
coil can return only a single number, so one coil can resolve
at most one degree of freedom of that pose. A typical
arrangement holds a conductive plate that fully covers the coil parallel to it and lets it translate only along the coil axis, so the one resolved degree of freedom is the standoff gap \cite{wang2018sits}. The remainder of this section recovers all six DOFs by using six coils and shaping the conductive plate, so that each pose component modulates the six couplings
differently.

\subsection{The Effect of Conductive Plate Geometry}
\label{sec:principle:paths}
\label{sec:principle:onedof}
\label{sec:design:whypattern}

A partially covered coil responds both to in-plane motion across the plate
edge and to changes in the gap \cite{wang2018triaxis}.
Each coil responds mainly to the part of the conductive plate within
a \emph{sensing zone} roughly the size of the coil. The conductor geometry
shapes the paths of the induced currents \cite{nagel2018}. At high
frequencies, the currents concentrate near the conductor surface
\cite[Sec.~2.2]{ti_target}. We therefore model the conductive plate as a
thin perfect electric conductor (PEC) sheet and compute its surface currents
with a numerical PEC solver. Fig.~\ref{fig:mech} shows the result for full
coverage and for a straight edge over the coil.
Under full coverage (Fig.~\ref{fig:mech}a), the current follows the coil outline.
Translating the plate leaves this path unchanged, so the coil is insensitive
to lateral motion. When a straight edge lies over the coil (b), the current follows the edge and the coil responds to motions in more than one direction.
Sliding the conductive plate then changes the inductance (d), with a tradeoff that the partial coverage reduces the sensitivity to
changes in gap (c). However, a single inductance measurement cannot distinguish
lateral motion from a change in gap.

\begin{figure}[!tb]
\centerline{\includegraphics[width=\columnwidth]{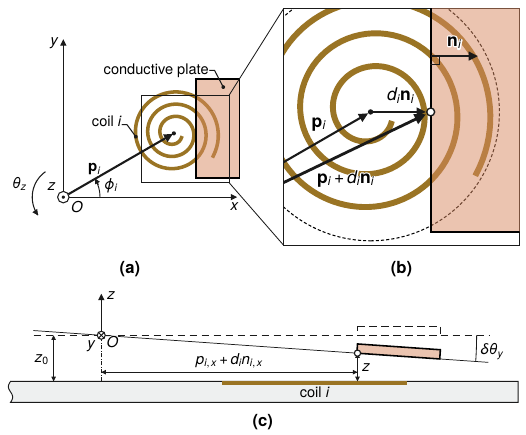}}
\caption{\textbf{Notation visualization for one coil under one edge.} (a) Coil $i$ at
$\mathbf{p}_i$ from the origin $O$, at the angle $\phi_i$ from the $x$
axis, under a straight edge of the conductive plate. (b) Enlarged
view. The edge normal $\mathbf{n}_i$ points into the conductive plate,
and the edge point nearest the coil center is
$\mathbf{p}_i+d_i\mathbf{n}_i$, at the edge offset $d_i$. (c) Side view
of a tilt $\delta\theta_y$. At the edge point the local gap changes from
the rest gap $z_0$ to $z$, in proportion to the lever arm
$p_{i,x}+d_i n_{i,x}$.}
\label{fig:notation}
\end{figure}

\subsection{The Sensor Jacobian and the Full-Rank Condition}
\label{sec:design:jacobian}

Six independent coil measurements are needed to resolve the three translations
and three rotations of the conductive plate. Let $\mathbf q=(x,y,z,\theta_x,\theta_y,\theta_z)$
denote the plate pose and $\mathbf q_0$ its rest pose under zero load. We call
rotations about the in-plane axes ($\theta_x$, $\theta_y$) \emph{tilts} and
rotation about the normal ($\theta_z$) a \emph{twist}. The
change in effective inductance of coil $i$ from rest is
$\Delta L_i=L_{\mathrm{eff},i}(\mathbf q)-L_{\mathrm{eff},i}(\mathbf q_0)$.
Stronger coupling than at rest therefore gives $\Delta L_i<0$.
For a small motion $\delta\mathbf q=\mathbf q-\mathbf q_0$, the six sensing
coil responses form the vector
\begin{equation}
    \Delta\mathbf L_{\mathrm{s}} \approx \mathbf S\,\delta\mathbf q,
    \qquad
    S_{ij}=\left.\frac{\partial L_{\mathrm{eff},i}}{\partial q_j}\right|_{\mathbf q_0},
    \label{eq:jacobian}
\end{equation}
where $\mathbf S$ is the $6\times6$ \emph{sensor
Jacobian}. Its columns $\mathbf{s}_j$ are the responses to the six
basis motions, each a pure translation along or rotation about one
axis. We call $\mathbf{s}_j$ the \emph{signature} of motion $j$, or the
signed inductance changes motion $j$ produces across the six coils. Any small
motion is a combination of the basis motions, so the inductance change
it produces is the combination of signatures,
$\Delta\mathbf L_{\mathrm{s}}\approx\sum_j \delta q_j\,\mathbf{s}_j$. If one
signature can be built from (is a linear combination of) the others, two different motions change
the six inductances in the same way, and the sensor cannot tell them
apart. Recovering the pose of the conductive plate relative to the coils
therefore requires $\mathbf{S}$ to have full rank.

\begin{figure*}[!tb]
\centerline{\includegraphics[width=\textwidth]{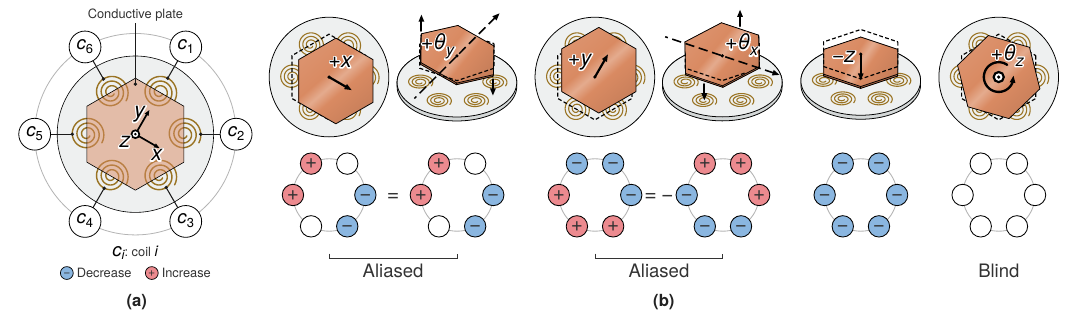}}
\caption{\textbf{Signatures of the hexagonal conductive plate.} Each of the six plate edges passes through a coil center, so every edge normal is radial. (a) Top view of the coils $c_1$ to $c_6$. (b) Each ring shows the effective inductance changes for the motion above it.
The downward press uses the negative of the $z$ column of $\mathbf S$ in
\eqref{eq:jacobian}. Blue ($-$) marks a decrease, red ($+$) an increase and
white a negligible response.
Dashed outlines depict the rest pose. For tilts, the long dashed line is the tilt axis, and the arrows show which
edges rise and which drop. Rings marked $=$ are aliased and $=-$ aliased with opposite signs. In the straight-edge model, $x$ is aliased with $\theta_y$,
$y$ with $\theta_x$, and $\theta_z$ is blind, so $\mathbf S$ has rank~$3$.}
\label{fig:patterning}
\end{figure*}

Rank is lost in two ways that the edge geometry makes visible. A
motion is \emph{blind} if its signature is negligible, so that it
changes no inductance. Two motions are \emph{aliased} if their
signatures are parallel, $\mathbf{s}_j = c\,\mathbf{s}_k$ for some $c \neq 0$, so the
coils report one motion as a scaled copy of the other. A negative $c$ makes it
a reversed copy, as in the pairs marked $=-$ in Fig.~\ref{fig:patterning}. Full rank determines whether all pose components
can be recovered. After scaling each column of $\mathbf{S}$ to its motion's working
half-range ($1.05$~mm in $x$ and $y$, $0.75$~mm in $z$, $0.95^\circ$ in
$\theta_x$ and $\theta_y$, and $3.9^\circ$ in $\theta_z$), the smallest singular
value $\sigma_{\min}$ of $\mathbf{S}$ is the smallest response magnitude to any
unit motion, and that motion is the weakest direction the sensor resolves.

A closed-form model of one coil under one straight edge makes the
entries of $\mathbf{S}$ explicit. Using the PEC solver of Sec.~\ref{sec:principle:paths} with a
single coil and a straight-edged conductive plate, we swept a lateral range spanning 0\% to 100\% coil coverage, and a vertical range of 1.2\,mm to 1.8\,mm around the 1.5\,mm rest gap. We fit the inductance reduction $M_i^2/L_{\mathrm{plate}}$ in \eqref{eq:leff}
to these data. The effective inductance is then
\begin{equation}
L_{\mathrm{eff},i}(d,z)
=
L_{\mathrm{coil},i}-\ell_0(d)\,
\frac{P(z)}{P(z_{0})},
\label{eq:response}
\end{equation}
where $z$ is the local gap, $d$ is the signed offset along the inward edge
normal (Fig.~\ref{fig:notation}b), $\ell_0(d)$ is the fitted reduction at
the rest gap $z_0$, and the gap factor $P(z)$ scales it with gap. The fitted
functions are
\begin{align}
\ell_0(d)
&=
\beta_{0}
+
\frac{\beta_{1}}
{1+\exp\!\bigl[-\beta_{2}(d-\beta_{3})\bigr]},
\label{eq:logistic}\\[4pt]
P(z)
&=
\beta_{4}^{-\beta_{5} z+\beta_{6}}.
\label{eq:exponential}
\end{align}
Increasing $d$ reduces coverage. The edge
passes through the coil center at $d=0$. For each coil radius and rest gap,
we fit $\beta_0,\ldots,\beta_3$ to the lateral sweep at $z_0$ and
$\beta_5$ to the gap sweep. We set $\beta_4=e$ and $\beta_6=\beta_5 z_0$,
so that $P(z_0)=1$.

Motion that is aligned with an edge does not change how much of the conductive plate overlaps the coil. If $\mathbf{n}_{i}$ is the unit edge normal, displacement along a plate edge is orthogonal to $\mathbf{n}_{i}$ and is therefore not observable by the coil.
Differentiating \eqref{eq:response} gives the two sensitivities that enter the sensor Jacobian,
\begin{align}
\frac{\partial L_{\mathrm{eff},i}}{\partial d}
&=
-\ell_0'(d)\,
\frac{P(z)}{P(z_{0})},
&
\ell_0'(d)
&=
\beta_{1}\beta_{2}\,\psi(d)\bigl(1-\psi(d)\bigr),
\label{eq:dfdr}\\[4pt]
\frac{\partial L_{\mathrm{eff},i}}{\partial z}
&=
-\ell_0(d)\,
\frac{P'(z)}{P(z_{0})},
&
P'(z)
&=
-\beta_{5}\ln(\beta_{4})\,P(z),
\label{eq:dfdd}
\end{align}
where $\psi(d)=\bigl(1+\exp[-\beta_{2}(d-\beta_{3})]\bigr)^{-1}$ is the logistic factor in \eqref{eq:logistic}.
Let coil $i$ sit at $\mathbf{p}_{i}=(p_{i,x},p_{i,y})$, with edge normal $\mathbf{n}_{i}$ and edge offset $d_{i}$. A small rigid displacement $\delta\mathbf{q}$ of the conductive plate changes the local edge offset and gap according to
\begin{align}
\delta d_{i}
&=
n_{i,x}\,\delta x
+
n_{i,y}\,\delta y
+
(p_{i,x} n_{i,y}-p_{i,y} n_{i,x})\,\delta\theta_{z},
\label{eq:dr}\\[4pt]
\delta z_{i}
&=
\delta z
+
(p_{i,y}+d_{i} n_{i,y})\,\delta\theta_{x}
-
(p_{i,x}+d_{i} n_{i,x})\,\delta\theta_{y}.
\label{eq:dd}
\end{align}
The tilt terms give the small-angle change in gap at the edge point nearest the
coil center, $\mathbf{p}_{i}+d_{i}\mathbf{n}_{i}$ (Fig.~\ref{fig:notation}b,c).
The chain rule then gives the $i$th row of the sensor Jacobian $\mathbf{S}$ of \eqref{eq:jacobian}, evaluated at the rest gap $z_{0}$,
\begin{equation}
\begin{aligned}
\mathbf{S}_{i}
=
\Big[
&
n_{i,x}\,\partial_d L_{\mathrm{eff},i},\;
n_{i,y}\,\partial_d L_{\mathrm{eff},i},\;
\partial_z L_{\mathrm{eff},i},\\
& (p_{i,y}+d_i n_{i,y})\,\partial_z L_{\mathrm{eff},i},\;
-(p_{i,x}+d_i n_{i,x})\,\partial_z L_{\mathrm{eff},i},\\
& (p_{i,x}n_{i,y}-p_{i,y}n_{i,x})\,\partial_d L_{\mathrm{eff},i}
\Big],
\end{aligned}
\label{eq:jacobian-row}
\end{equation}
with both derivatives evaluated at $(d_i,z_0)$ using \eqref{eq:dfdr} and
\eqref{eq:dfdd}. Stacking the six rows gives $\mathbf S$. The Jacobian of
the inductance reduction is $-\mathbf S$ and has the same singular values.

Analyzing the generic Jacobian row, $\textbf{S}_i$, is informative to the effects of modifying coil layout and plate geometry. Coils which are further from the measurement point (larger $||\textbf{p}_i||$) provide more useful angular information due to a longer lever arm. There exists a direct tradeoff between sensitivity in $x$, $y$, and $\theta_z$, which is determined primarily by the edge normal $\textbf{n}_i$. The edge offset $d_i$ affects the local sensitivities to the gap and edge offset, and has a small effect on the $\theta_x$ and $\theta_y$ sensitivity by extending the lever arm. If a singular Jacobian is formed by the conductive plate, it is non-invertible and cannot resolve all poses uniquely.

Figs.~\ref{fig:patterning} and~\ref{fig:signtable} show examples of conductive plates with different edge normals, but identical coil positions. A hexagonal conductive plate (Fig.~\ref{fig:patterning}) has radial normals, $\mathbf{n}_i=-\mathbf{p}_i/\lVert\mathbf{p}_i\rVert$, so $p_{i,x}n_{i,y}-p_{i,y}n_{i,x}=0$, rendering $\theta_z$ blind. In the straight-edge model, the $x$ and $\theta_y$ signatures are proportional, as are the $y$ and $\theta_x$ signatures. This leaves three independent signatures.
The first square target of Zhang \emph{et al.} aliases the same two pairs
\cite{zhang2024rmfs}.

\subsection{The Three-Blade Conductive Plate and Its Sign Table}
\label{sec:design:pattern}

\begin{figure*}[!tb]
\centerline{\includegraphics[width=\textwidth]{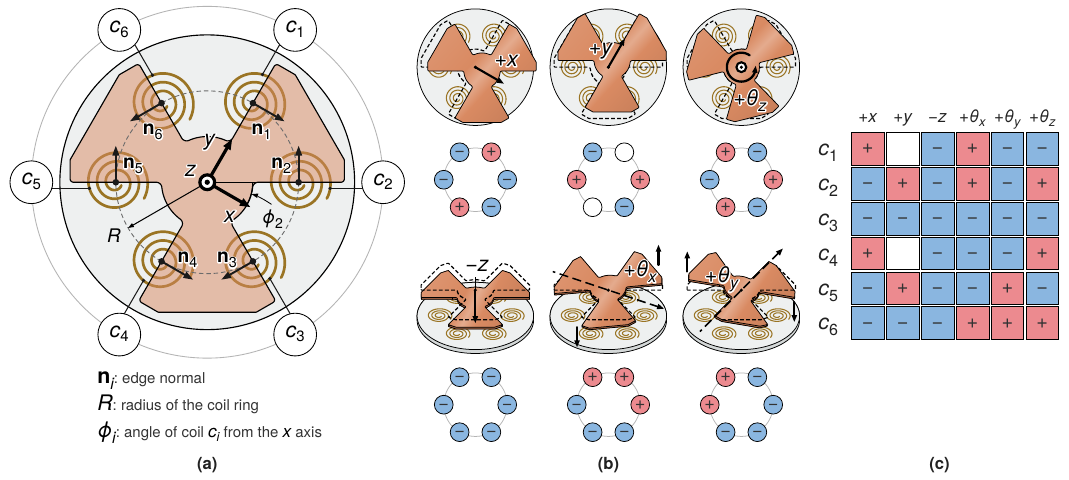}}
\caption{
\textbf{Signatures of the initial three-blade conductive plate.} Rest gap $1.5$~mm. (a) Three blades on a hub give six
radial edges, one through each coil center. Their edge normals turn alternately clockwise and counterclockwise. (b) The six motions and their signatures, colored as in Fig.~\ref{fig:patterning}. (c) Signs of the effective inductance changes from the PEC model, with one
row per coil and one column per motion. The press column shows $-z$. For small motions relative to the rest pose, $\mathbf{S}$ has rank $6$. Scaling each column by its motion's working half-range gives $\sigma_{\min}=30$~nH and a condition number of $8$.}
\label{fig:signtable}
\end{figure*}

The initial plate in Fig.~\ref{fig:signtable}a is the simplest design
considered here whose Jacobian has full rank. Six radial cuts leave three
blades joined at a central hub, with one edge crossing each coil center.
The inward edge normals point alternately clockwise and counterclockwise
around the ring. In the model, opposite coils respond equally to a lateral
translation because their normals point the same way. A tilt gives opposite
responses within each pair. A twist produces alternating signs around the
ring, while a gap change gives the same sign at every coil (Fig.~\ref{fig:signtable}b).

At the initial pose ($d_i=0$, $z=z_0$), both $\partial_d L_{\mathrm{eff},i}$
and $\partial_z L_{\mathrm{eff},i}$ are positive. Thus in
\eqref{eq:jacobian-row}, with each normal
pointing into the conductive plate, the $x$ and $y$ columns take the signs of
$n_{i,x}$ and $n_{i,y}$, and the $\theta_z$ column takes the sign of
$p_{i,x}n_{i,y}-p_{i,y}n_{i,x}$, which alternates from coil to coil.
The $+z$ response is positive at every coil, so a downward press lowers all
six inductances. The $\theta_x$ and
$\theta_y$ columns follow $p_{i,y}$ and $-p_{i,x}$, the place of each
coil on the ring. The six motions have distinct sign patterns, with at least
four coils responding to each lateral translation or twist.
Fig.~\ref{fig:signtable}c shows the signs from the full PEC model.
The straight-edge model, which the optimization in Sec.~\ref{sec:opt} uses,
predicts the same sign for 34 of the 36 coil--motion pairs. The two exceptions
are coils $c_1$ and $c_4$ under a positive tilt about $y$, where the
straight-edge model predicts no response and the full PEC model a small
decrease, about a tenth of the other $\theta_y$ responses.

Full rank is necessary but not sufficient. The tilt columns are the
weakest, by a factor of five to eight, and it is they that set
$\sigma_{\min}$. The design fixes the topology but not the exact angle of each edge
or where the edge crosses its coil. Those choices set $\sigma_{\min}$, and
Sec.~\ref{sec:opt} determines them by optimizing the edges of the conductive plate. 

\subsection{Optimizing the Conductive Plate}
\label{sec:opt}

We developed a flexible optimization approach to
determine the optimal shape of the conductive plate for a generic set of coils.
The method described in Sec.~\ref{sec:design:jacobian} yields $\mathbf{S}$ for any
candidate conductive plate at negligible computational cost, making it inexpensive to use in an optimizer.

The silicone pillars support the conductive plate above the coil board
(Fig.~\ref{fig:teaser}). Near the rest pose, their compliance relates a small
change in contact wrench to plate motion, $\delta\mathbf{q}\approx\mathbf{C}\,\delta\mathbf{W}$.
We estimated the diagonal compliance matrix $\mathbf{C}$ for the optimization
using finite-element analysis in Abaqus. Combining this relation
with \eqref{eq:jacobian} gives
\begin{equation}
    \Delta\mathbf L_{\mathrm{s}} \approx \mathbf{J}_{W}\,\delta\mathbf{W},
    \qquad \mathbf{J}_{W}=\mathbf{S}\,\mathbf{C},
    \label{eq:forcejac}
\end{equation}
where $\mathbf{J}_{W}$ maps wrench changes to inductance changes. The pillars
are stiffer in compression than in shear (Sec.~\ref{sec:build:fab}).

The diagonal matrix $\boldsymbol{\Lambda}$ scales the force and moment
columns of $\mathbf{J}_W$ by their chosen load ranges. We reduced the shear
weights a hundredfold to favor shear sensitivity in the optimization.

Larger singular values of $\mathbf{J}_{W}\boldsymbol{\Lambda}$ produce greater
changes in inductance for the same change in the wrench, and the minimum
singular value indicates the sensor's weakest direction.

Searches with six independent edges gave nearly symmetric layouts. For the
final design, we vary the unit normal $\mathbf{n}_0$ and signed offset $d_0$
of one edge. Reflection across the blade centerline gives the other edge,
and rotations about the plate center by $120^\circ$ and $240^\circ$ give the other two blades.
We maximize the minimum
singular value of $\mathbf{J}_{W}\boldsymbol{\Lambda}$. A small regularization
term is added, equal to a scaled generalized mean of order
$-2$ of the singular values, to differentiate between solutions
with nearly equal minimum singular values while ensuring that weaker singular values are prioritized:
\begin{equation}
\begin{aligned}
\max_{\mathbf{n}_{0},d_{0}}
\;
J
&=
\sigma_{\min}\!\bigl(\mathbf{J}_{W}\boldsymbol{\Lambda}\bigr)
+
\eta\,
M_{\sminus2}\!\bigl(\sigma(\mathbf{J}_{W}\boldsymbol{\Lambda})\bigr),
\\[6pt]
M_{\sminus2}(\sigma)
&=
\Biggl(
\frac{1}{m}
\sum_{i=1}^{m}
\sigma_{i}^{-2}
\Biggr)^{-1/2},
\end{aligned}
\label{eq:cost}
\end{equation}
where \(\sigma_{1}\ge\cdots\ge\sigma_{m}\) are the singular values of
\(\mathbf{J}_{W}\boldsymbol{\Lambda}\),
\(\sigma_{\min}=\sigma_{m}\), \(\eta>0\) is the regularization weight, and
\(M_{\sminus2}\) is taken to be zero if any \(\sigma_{i}=0\).

\begin{figure}[!tb]
\centerline{\includegraphics[width=\columnwidth]{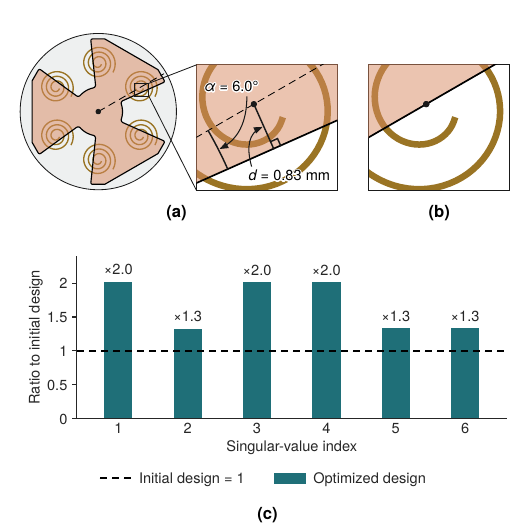}}
\caption{\textbf{Optimized three-blade conductive plate.} (a) Layout and enlarged view of one coil. The edge makes an angle $\alpha$ with the dashed line between the plate and coil centers. The dimension labeled $d$ gives the offset magnitude $|d_i|$.
(b) Enlarged view of the initial design (Fig.~\ref{fig:signtable}) whose edge passes through the coil center, for comparison. (c) Singular values of $\mathbf{J}_{W}\boldsymbol{\Lambda}$ from \eqref{eq:forcejac}, largest first, relative to those of the initial design.}
\label{fig:optimization}
\end{figure}

We optimized the symmetric design with the Nelder--Mead method~\cite{nelder1965}
from 1,000
random starts. Allowing the two edges of the reference blade to vary independently
gave the same gains to the precision shown in Fig.~\ref{fig:optimization}c.

Across the directional weights tested, the optimized three-blade designs had
$67\%$ to $98\%$ coil coverage. Greater coverage increases
out-of-plane sensitivity, while coverage near $67\%$ favors in-plane sensitivity.

Fig.~\ref{fig:optimization} compares the initial design with the one optimized for near-maximal shear sensitivity. The optimizer maintains the three-blade topology and
widens each blade, turning its two edges $6^\circ$ outward about the
center of the conductive plate (Fig.~\ref{fig:optimization}a,b).
The eddy currents screen the coil's axial field within the plate and
reinforce it just outside the edge. As the edge advances past the coil
center, the plate diverts more of the strong central flux around the edge.
Each additional strip of conductor then screens a stronger field, increasing
the change in coil flux linkage per unit displacement (Fig.~\ref{fig:mech}d).
Farther out, the field at
the edge weakens and lateral sensitivity falls. At the $1.5$~mm rest gap,
the fitted response in \eqref{eq:logistic} has its largest slope magnitude at $d_i=-0.83$~mm, where the plate covers
$67\%$ of the coil.
The shear forces and the twisting moment therefore
gain a factor of $1.3$. Normal force and the two tilting moments gain
a factor of $2.0$, because more of the conductive plate over each coil strengthens the
gap channel through which they act. The edges turn rather than shift
because an edge shifted parallel to itself would couple each shear
force to a tilting moment, while an edge through the center of the conductive plate
keeps those columns orthogonal. The conductive plate of
Fig.~\ref{fig:patterning} lost three singular values entirely.
The initial three-blade design retains all six,
and optimization increases each singular value, with a $1.3$-fold gain in the
smallest (Fig.~\ref{fig:optimization}c). The fabrication drawing for the
sensor in Sec.~\ref{sec:build} uses this optimized edge layout.

\FloatBarrier
\section{Sensor Construction}
\label{sec:build}

\begin{figure*}[!tb]
\centerline{\includegraphics[width=\textwidth]{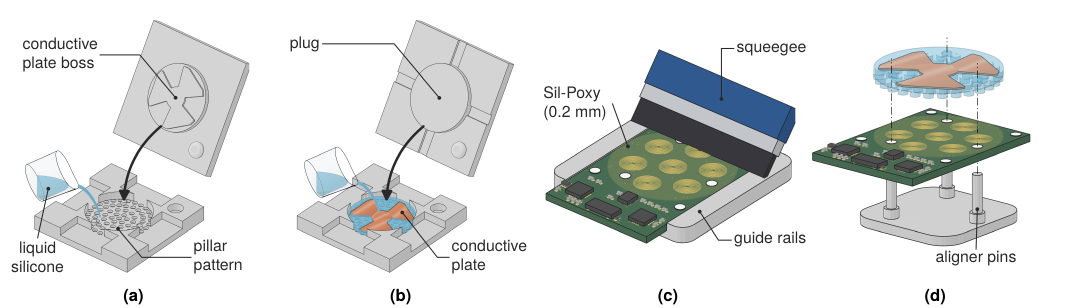}}
\caption{
\textbf{Sensor fabrication process.} (a) Liquid silicone poured into the mold
base fills the pillar pattern and forms the pillared silicone. The boss on the
first lid leaves a pocket in the shape of the conductive plate. (b) The
conductive plate sits in the pocket, and a second pour forms the cover silicone
over it. The plug on the second lid flattens the contact surface. (c) A
squeegee running on guide rails spreads a $0.2$~mm layer of silicone adhesive
(Sil-Poxy) on the coil board. (d) Pins on the aligner piece position the cured
silicone stack over the coils as it is placed (exploded view).}
\label{fig:fabrication}
\end{figure*}

\subsection{Coil Board and Conductive Plate}
\label{sec:build:board}

The coil board (Fig.~\ref{fig:teaser}) is a four-layer rigid PCB,
$0.8$~mm thick, fabricated and assembled by JLCPCB. Its $25.8\times36.6$~mm
footprint fits a finger of the Universal Manipulation Interface (UMI)
gripper~\cite{umi}. The six sensing coils
lie on a $16$~mm diameter circle, with an auxiliary coil at its center.
Each coil is nominally $6$~mm in diameter and has $11$ turns per layer.
The four layers are wound in the same direction and connected in series.
The conductive plate is cut from $0.3$~mm copper sheet to the three-blade
outline of Sec.~\ref{sec:opt}, with one edge over each sensing coil. The
silicone stack covers only the sensing area. The readout electronics
occupy a separate strip on the same board. The BOM cost for EddyTac is
\$9.21 per sensor at 1,000 units.

\subsection{Resonant Readout}
\label{sec:build:readout}
\label{sec:pose:setup}

Each coil of the array is paired with a fixed capacitor into a
resonant tank whose frequency
\begin{equation}
    f_i = \frac{1}{2\pi\sqrt{L_{\mathrm{eff},i}\,C_i}}, \qquad
    i = 1, \dots, 7.
    \label{eq:tank}
\end{equation}
is read by a multichannel inductance-to-digital converter. As the
conductive plate approaches a coil, $L_{\mathrm{eff},i}$ falls and $f_i$
rises. Two four-channel LDC1614 converters \cite{ldc1614} read the
seven coils. The six sensing coils already need both converters, so the
auxiliary coil takes a spare channel and needs no extra hardware. Each converter excites its channels sequentially, and the two converters run in parallel, so at any
moment each converter drives only one tank. Two tanks driven together at the same
frequency would interfere and add noise, so the capacitors (C0G
ceramics, $\pm 1\%$) differ
by converter rather than by coil ($100$~pF on every channel of the
first converter and $150$~pF on every channel of the second). With the designed coil inductance of $5.1$~$\mu$H,
these capacitors place the first converter's tanks at $7.1$~MHz and the
second's at $5.8$~MHz, $22\%$ apart. Measured on an EddyTac sensor with
the conductive plate in place and no load, the resonances are
$6.85$--$6.94$~MHz and $5.70$--$5.94$~MHz, so any two tanks driven
together are at least $15\%$ apart. Each converter steps through its
four channels, $1.66$~ms each, so every coil is sampled at $150$~Hz.
The converters sit in the readout strip with a digital temperature
sensor and reference oscillator. A flat flexible cable (FFC) carries I$^2$C bus and power
to a separate microcontroller readout board. The sensing electronics therefore add no
thickness or stiffness to the sensing area, and only digital signals
leave the coil board, lowering noise. Since the analog circuitry stays on the
coil board, the FFC can be any length the mounting needs.

\subsection{Silicone Stack}
\label{sec:build:fab}
\label{sec:device:fab}
\label{sec:device:stack}

The silicone stack is cast from a Shore 10A platinum-cure silicone
(Dragon Skin 10 NV, Smooth-On) in a printed mold with a base and two lids
(Fig.~\ref{fig:fabrication}). The first pour fills the base, whose pillar
pattern forms the pillared silicone that sets the rest gap. The first lid
then closes the mold, and a boss cut to the outline of the conductive
plate leaves a shallow pocket in the top of this layer. Once the layer has
cured, the conductive plate is placed in the pocket to align it. The
second pour fills the pocket over the plate with the cover silicone, and the
second lid closes the mold with a flat plug that forms the contact surface.
Each pour is degassed in a vacuum chamber to remove bubbles and cures at room
temperature for $75$~minutes, so the process needs only common lab or
makerspace equipment.

The released stack is bonded to the coil board with silicone adhesive
(Sil-Poxy, Smooth-On). A squeegee running on guide rails spreads a
$0.2$~mm layer on the board. An aligner with pins positions the stack over
the coils as it is placed and is then removed. The adhesive reaches full
cure in $24$~h per manufacturer recommendation. The mold
and the aligner are the only alignment steps, with no fine alignment required. The calibration of each sensor
absorbs small placement errors.

The stack can be changed independently of the coil board and
conductive plate. For a fixed geometry, a softer silicone allows greater plate displacement
under the same load to increase sensitivity, but narrows the load
range.

The geometry of the pillared silicone sets the ratio between the lateral and compressive sensitivities. The compressive behavior of an incompressible rubber pillar bonded between two surfaces depends on its shape factor $s$, and its effective compressive modulus grows with $s$ as \cite{gentModel}:

\begin{equation}
    E_c = 3G\left(1 + 2s^2\right),
    \label{eq:shape_factor}
\end{equation}
where $E_c$ is the effective compressive modulus, and $s=\frac{r}{2t}$ is the shape factor defined by the ratio between pillar radius, $r$, and height, $t$. With $N$ pillars in parallel, the pillar radius sets the shape factor and $N$ sets the overall stiffness. This provides the stiffness under shear and compression, and their ratio:

\begin{equation}
    \begin{aligned}
        K_c &= \frac{E_c A N}{t}, \qquad
        K_s &= \frac{G A N}{t}.
    \end{aligned}
    \label{eq:stiffness}
\end{equation}

\begin{equation}
     \frac{K_c}{K_s}
    = \frac{E_c}{G}
    = 3\left(1+2s^2\right).
    \label{eq:stiffness_ratio}
\end{equation}
where $N$ is the number of pillars in parallel, $G$ is the shear modulus of the material, $K_c$ and $K_s$ are the compressive and shear stiffness, and $A$ is cross-sectional area. The incorporation of multiple pillars is crucial, because a thin bonded layer of nearly incompressible silicone is very stiff in compression: if $r=12$~mm and $t=1$~mm, $s=6$ resulting in $K_c/K_s=219$. In our stack, the 58 pillars are $2$~mm in diameter and $1$~mm tall, which gives $s=0.5$ and $K_c/K_s=4.5$.

\section{Force--Torque Calibration and Evaluation}
\label{sec:device}

We calibrate one EddyTac sensor against a reference F/T sensor and
evaluate it in a separate test session. This section details the calibration method and also examines the contribution
of the auxiliary coil.

\begin{figure}[!tb]
\centerline{\includegraphics[width=\columnwidth]{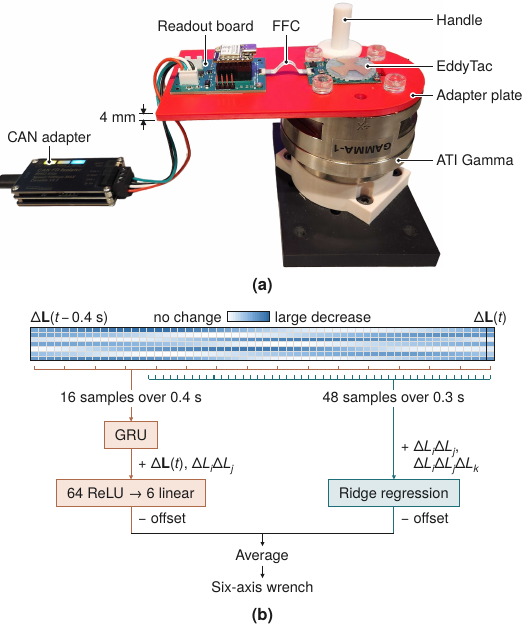}}
\caption{
\textbf{Calibration setup and wrench estimation.} (a) The adapter plate connects
the tactile sensor to the reference sensor, and the operator applies loads by
hand through the handle. (b) The calibration model. $\Delta\mathbf{L}(t)$ holds the inductance
changes of the seven coils from tare, and $t$ is the time of the latest scan.
The values are illustrative, and the ticks mark the samples each branch
reads. $\Delta L_i\Delta L_j$ and $\Delta L_i\Delta L_j\Delta L_k$ are products of the
standardized latest readings, taken over every combination of the seven coils
with $i\le j\le k$, squares and cubes included. Each branch's offset is its
output when every coil input is held at its tare value.}
\label{fig:calibrig}
\end{figure}

\begin{figure*}[!tb]
\centerline{\includegraphics[width=\textwidth]{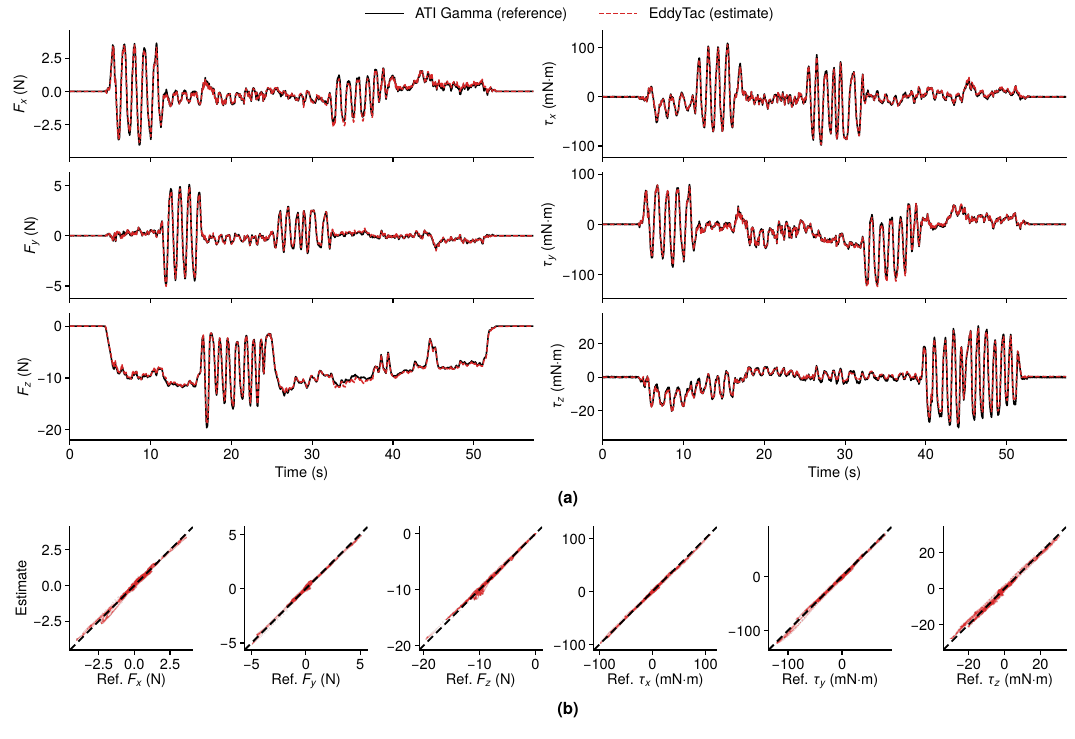}}
\caption{
\textbf{Wrench estimates during the test session.}
(a) Real-time force and moment estimates over time with the reference measurements.
(b) Estimates against the reference for the same samples.
The dashed line marks exact agreement. Table~\ref{tab:ft_accuracy} lists the errors.}
\label{fig:ftresult}
\label{fig:ftparity}
\end{figure*}

\subsection{Calibration Model}
\label{sec:device:wrench}

The model estimates three forces and three moments from changes in the
seven coil inductances relative to tare. It combines two branches, a ridge regression and a
gated recurrent unit (GRU) network (Fig.~\ref{fig:calibrig}b).
For each branch, we compute a wrench offset by holding every coil input at
its tare value throughout the input window. We subtract each branch's offset
from its predicted wrench and average the two corrected estimates.

The linear terms in the ridge regression use 48 evenly spaced samples from
the past 0.3~s without clipping. To form the quadratic and cubic terms, we
clip each coil's latest reading to its training range before multiplication.
A 16-unit GRU processes 16 evenly spaced samples
from the past 0.4~s. Its final state, the latest readings, and their pairwise products feed a 64-unit ReLU layer followed by six linear outputs.

\textbf{Product terms.} The coils' response to a load
changes with the load itself. In linear fits of the coil responses within
bands of normal force, the sensitivities to $F_z$, $\tau_x$, and $\tau_y$ fall
by 26--36\% from 1--4~N to 14--25~N, while those to
$F_x$, $F_y$, and $\tau_z$ rise by 7--28\%.
These trends are consistent with silicone stiffening and a smaller plate-to-coil gap under compression.
A shear estimate therefore has to depend on the coils
that register the press. Products of pairs of readings provide that
dependence, and squares and cubes follow the falling normal-force
sensitivity. Pairwise products cut the shear-force and tilting-moment errors
of a linear ridge regression by 39--55\%, and triple products cut the $F_z$
error by a further 27\% (Table~\ref{tab:ablation}).

\textbf{Two branches.} The GRU network was more accurate than the ridge
regression in shear and twist, particularly for contacts more than 6~mm from
the center of the sensing area, and the ridge regression was more accurate
in $F_z$ at loads of 3--10~N. The squared error of an average equals the mean error
of its members minus their spread around the average~\cite{krogh1995}, so
averaging helps most when the members err in different places. The average
had the lowest error on every axis (Table~\ref{tab:ablation}) and a higher
six-axis mean $R^2$ than either branch alone in all six leave-one-session-out
folds. Wang and Liu instead assign polynomial regression to the forces and a
neural network to the moments of a six-axis force sensor~\cite{wang2024decoupling}.
In our model search, larger networks and inputs that summarize a longer
history or model hysteresis did not consistently improve accuracy.

The six training sessions provided 13.2~min of data.
We standardized the coil inputs and reference wrenches, fitted the ridge
coefficients by regularized least squares, and trained the GRU network
with Adam using Huber loss. For the GRU, we selected the number of epochs
on the last 20\% of each training session, leaving a 5~s gap before
validation, then refit on the full training data. We evaluated the final
model on 57.4~s of data from a separate test session.
The complete model has 7,612 parameters. Median inference time was
0.7~ms on one CPU thread, within the 6.7~ms scan interval.

\subsection{Experimental Setup and Protocol}
\label{sec:device:calib}

The sensor is mounted on an ATI Gamma reference F/T sensor through a 3D-printed adapter plate (Fig.~\ref{fig:calibrig}a). An operator loads it by
hand through a handle whose flat base rests on the contact surface. Since
the handle only rests on the surface, every load was applied while pressing
($F_z<0$). The two converters scan all seven coils at $150$~Hz. Each scan
was paired with the latest ATI reading, which arrives at about 7~kHz.

\textbf{Reference frame.} We report moments about the ATI Gamma origin
at the center of its tool face. At a point $\mathrm{S}$ displaced by $r_z$ toward the contact surface, the
estimated moments are
\begin{equation}
    \hat{\tau}_{\mathrm{S},x} = \hat{\tau}_{x} + r_z\hat{F}_y,
    \qquad
    \hat{\tau}_{\mathrm{S},y} = \hat{\tau}_{y} - r_z\hat{F}_x,
    \label{eq:frame_transfer}
\end{equation}
while $\hat{\tau}_{z}$ is unchanged.

\textbf{Loading protocol.} One 300~s guided session included normal
presses, lateral pushes along both axes, off-center presses, and twists,
followed by free combinations. In five additional sessions, the operator
watched the real-time estimate and applied loads where the estimate was poor.
Training loads reached 34.5~N in compression (Table~\ref{tab:ft_accuracy}).
The test loads covered 41--63\% of the training load span, depending on
the axis. Peak test loads were 19.6~N in compression, 5.1~N in resultant
shear, and 122~mN$\cdot$m in resultant moment. Overload capacity was not tested.

\textbf{Preprocessing.} For the real-time estimates, each coil was tared once at
startup to the mean of 20 unloaded scans. For training and offline estimates, each coil's baseline
was its mean from 1 to 3~s.

\textbf{Evaluation.} We express mean and maximum absolute errors as a
percentage of each axis's training load span (\%FS). For leave-one-session-out
(LOSO) validation, we fit the model on five training sessions with early
stopping and evaluated it on the sixth, rotating through all six sessions.

\subsection{Six-Axis Estimation Accuracy}
\label{sec:device:accuracy}

During the test, the operator applied different combinations of forces
and moments by hand. The RMSE of the real-time estimates was 0.09--0.26~N
for forces and 1.0--2.1~mN$\cdot$m for moments
(Fig.~\ref{fig:ftresult}, Table~\ref{tab:ft_accuracy}).

Across the six LOSO folds, mean RMSE in $x$, $y$, and $z$ was
0.172, 0.149, and 0.354~N for forces and 3.49, 3.31, and
1.57~mN$\cdot$m for moments (Table~\ref{tab:ablation}). The six-axis mean
$R^2$ was $0.990\pm0.003$ (mean $\pm$ SD across folds). These are means over
held-out sessions whose accuracy varied, with $F_z$ RMSE ranging from 0.23 to
0.51~N. The test-session errors in Table~\ref{tab:ft_accuracy} are close to, or
lower than, those of the most accurate held-out sessions. The deployed model
was trained on all six sessions, whereas each fold's model was trained on five.

\textbf{Auxiliary coil.} A six-coil model retrained without the auxiliary
coil had 5--9\% higher moment RMSE on the test session, and 0--8\% lower force
RMSE (Table~\ref{tab:ft_coils}). The auxiliary coil uses a spare converter
channel (Sec.~\ref{sec:build:readout}), so it improves the moment estimates
without an additional inductance converter.

\begin{table}[!tb]
\caption{
Wrench Estimation Accuracy}
\label{tab:ft_accuracy}
\centering
\setlength{\tabcolsep}{3pt}\footnotesize
\sbox{\tabnotebox}{%
\begin{tabular}{lcccccc}
\hline
 & & & \multicolumn{2}{c}{Error (\%FS)} & & \\
\cline{4-5}
Axis & Training range & Test range & Mean & Max. & RMSE & $R^2$ \\
\hline
$F_x$ (N) & $-9.3$--$6.6$ & $-4.0$--$3.6$ & 0.49 & 2.91 & 0.110 & 0.989 \\
$F_y$ (N) & $-8.7$--$7.1$ & $-4.9$--$5.1$ & 0.39 & 2.62 & 0.089 & 0.994 \\
$F_z$ (N) & $-34.5$--$0.05$ & $-19.6$--$0.03$ & 0.53 & 3.91 & 0.265 & 0.996 \\
$\tau_{x}$ (mN$\cdot$m) & $-177$--$208$ & $-98$--$109$ & 0.33 & 1.96 & 1.70 & 0.996 \\
$\tau_{y}$ (mN$\cdot$m) & $-187$--$155$ & $-121$--$79$ & 0.40 & 3.34 & 2.09 & 0.995 \\
$\tau_{z}$ (mN$\cdot$m) & $-87$--$60$ & $-30$--$31$ & 0.50 & 2.67 & 1.02 & 0.988 \\
\hline
\end{tabular}}
\usebox{\tabnotebox}
\tabnote{Accuracy of the real-time wrench estimates in the test session. Full scale (FS) is the training
load span on each axis.}
\end{table}

\begin{table}[!tb]
\caption{
Six- versus Seven-Channel Calibration}
\label{tab:ft_coils}
\centering
\setlength{\tabcolsep}{2pt}\footnotesize
\sbox{\tabnotebox}{%
\begin{tabular}{llccccccc}
\hline
Coils & Error & $F_x$ & $F_y$ & $F_z$ & $\tau_{x}$ & $\tau_{y}$ & $\tau_{z}$ & $R^2$ \\
\hline
$6$ sensing & Mean \%FS & 0.48 & 0.36 & 0.51 & 0.33 & 0.41 & 0.53 & 0.993 \\
 & Max.\ \%FS & 2.98 & 2.19 & 3.88 & 2.45 & 4.59 & 3.04 &  \\
 & RMSE & 0.106 & 0.083 & 0.268 & 1.80 & 2.20 & 1.12 &  \\
$6{+}1$ aux. & Mean \%FS & 0.49 & 0.38 & 0.53 & 0.33 & 0.40 & 0.50 & 0.993 \\
 & Max.\ \%FS & 2.92 & 2.39 & 4.08 & 1.92 & 3.36 & 2.76 &  \\
 & RMSE & 0.108 & 0.090 & 0.269 & 1.71 & 2.07 & 1.03 &  \\
\hline
\end{tabular}}
\usebox{\tabnotebox}
\tabnote{Both models were retrained using the same procedure and evaluated offline on the same test scans. Units and full scales follow
Table~\ref{tab:ft_accuracy}. $R^2$ is the six-axis mean.}
\end{table}

\begin{table}[!tb]
\caption{
Calibration Model Ablation}
\label{tab:ablation}
\centering
\setlength{\tabcolsep}{2.5pt}\footnotesize
\sbox{\tabnotebox}{%
\begin{tabular}{lccccccc}
\hline
 & \multicolumn{6}{c}{RMSE} & \\
\cline{2-7}
Model & $F_x$ & $F_y$ & $F_z$ & $\tau_{x}$ & $\tau_{y}$ & $\tau_{z}$ & $R^2$ \\
\hline
Linear ridge & 0.396 & 0.302 & 0.671 & 6.56 & 9.05 & 2.52 & 0.954 \\
\quad + pair products & 0.219 & 0.183 & 0.541 & 3.89 & 4.09 & 1.99 & 0.984 \\
\quad + triple products & 0.202 & 0.168 & 0.395 & 3.87 & 3.69 & 1.80 & 0.987 \\
GRU branch & 0.177 & 0.161 & 0.403 & 3.73 & 3.67 & 1.67 & 0.988 \\
Average of both branches & \textbf{0.172} & \textbf{0.149} & \textbf{0.354} & \textbf{3.49} & \textbf{3.31} & \textbf{1.57} & \textbf{0.990} \\
\hline
\end{tabular}}
\usebox{\tabnotebox}
\tabnote{Means over the six leave-one-session-out folds, in N and mN$\cdot$m.
Each fold retrains every model on the other five sessions.
Each indented row adds terms to the ridge regression above it, and the
third row is the ridge branch. Pair products include squares and triple
products include cubes, all of the latest readings. $R^2$ is the six-axis
mean.}
\end{table}

\FloatBarrier
\section{Applications}
\label{sec:applications}

\begin{figure}[!tb]
\centering
\includegraphics[width=\columnwidth]{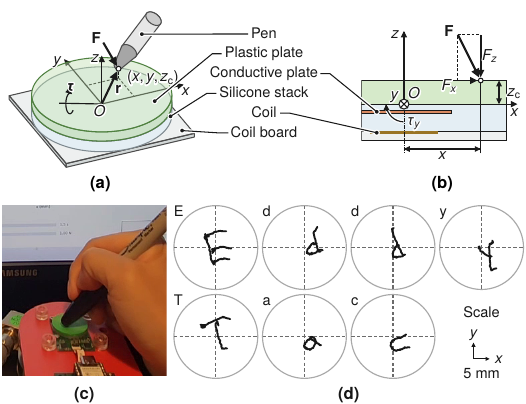}
\caption{\textbf{Handwriting demonstration on the sensor.}
(a) The pen applies force $\mathbf{F}$ at $\mathbf{r}=(x,y,z_{\mathrm{c}})$,
giving a moment $\boldsymbol{\tau}=\mathbf{r}\times\mathbf{F}$ about $O$ at
the center of the cover silicone's top face. (b) In the $xz$ section through
$O$, $\tau_y=z_{\mathrm{c}}F_x-xF_z$. (c) Writing on the plate with a
pen. (d) Estimated paths of seven letters at the same scale. Dashed lines
mark the sensor axes through the center of the silicone stack.}
\label{fig:handwriting}
\end{figure}

\begin{figure}[!tb]
\centerline{\includegraphics[width=\columnwidth]{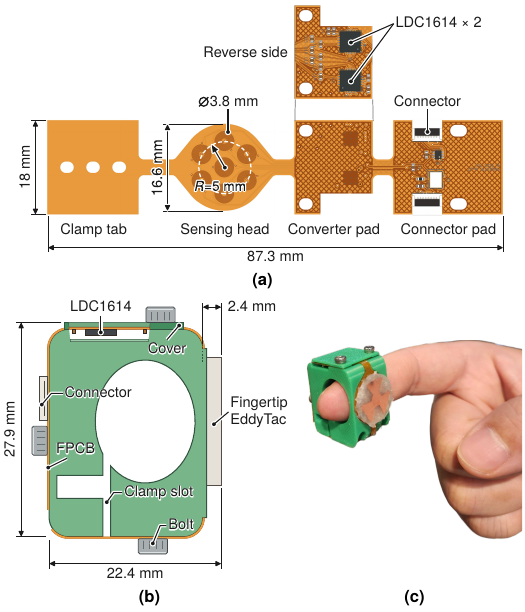}}
\caption{\textbf{Fingertip version of EddyTac.} (a) The FPCB laid flat.
The inset shows the back of the pad carrying the electronics.
$R$ is the radius of the sensing-coil ring. (b) The same board wrapped
around the thumb mount. The cutaway shows a converter beneath the side
cover. The overall dimensions exclude the bolt heads. The $2.4$~mm thickness includes the FPCB. (c) The sensor worn on the index finger.}
\label{fig:fingertip}
\end{figure}

\begin{figure*}[!t]
\centering
\includegraphics[width=\textwidth]{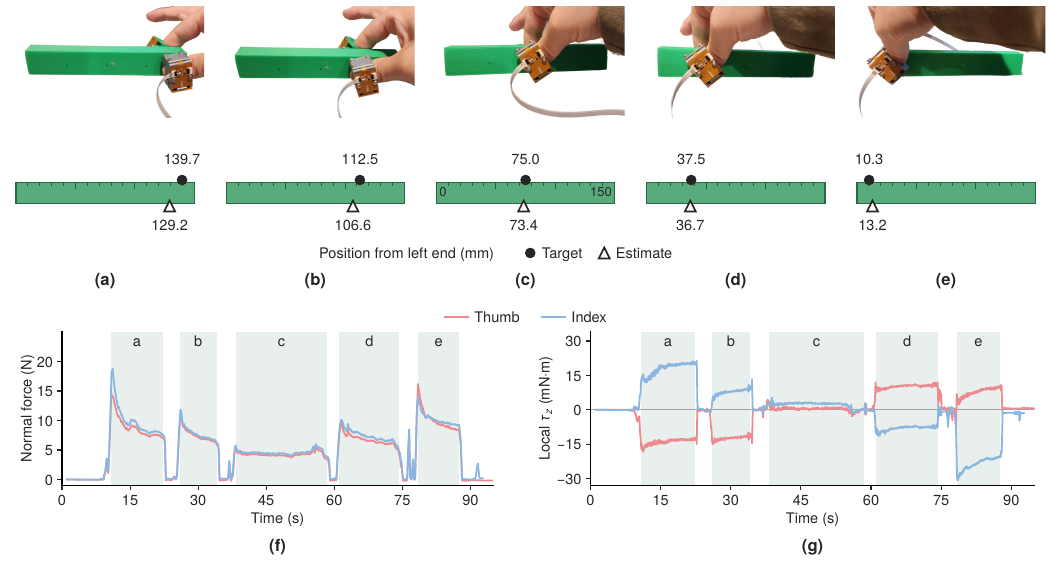}
\caption{\textbf{Demonstrating the fingertip sensor for grasping positions along a bar.} (a)--(e) Five successive grasps
from right to left. Circles mark the intended grasp centers and triangles
mark the estimates, both given as distances from the left end. Short ticks
mark $10$~mm intervals and long ticks $50$~mm. (f) Normal force at the thumb
and index finger. (g) Moment about each sensor's local $z$ axis. Shading
marks the steady interval used to estimate each grasp position.}
\label{fig:pinch}
\end{figure*}

\subsection{Locating the Center of Pressure}

The estimated wrench can also give the CoP of a contact on a surface of known
height. To test how well the sensor tracks the CoP, an operator wrote the
letters of \textit{EddyTac} with a pen on a calibrated
EddyTac sensor, covered by a plastic plate whose top surface lies at height $z_{\mathrm{c}}=2$~mm above
the cover silicone (Fig.~\ref{fig:handwriting}c). We move the moments by
\eqref{eq:frame_transfer} to $O$, the center of the top face of the cover
silicone (Fig.~\ref{fig:handwriting}a, b). The 4~mm thick adapter plate and the 3.1~mm thick
sensor place $O$ at $r_z=7.1$~mm above the ATI origin. On a contact surface at
height $z_{\mathrm{c}}$ above $O$, the CoP
is then
\begin{equation}
  \hat{x} = \frac{z_{\mathrm{c}} \hat{F}_x - \hat{\tau}_y}{\hat{F}_z},
  \qquad
  \hat{y} = \frac{\hat{\tau}_x + z_{\mathrm{c}} \hat{F}_y}{\hat{F}_z}.
  \label{eq:contact_position}
\end{equation}
The terms in $z_{\mathrm{c}}$ account for the moment of the tangential
force, which acts at that height. Equation~\eqref{eq:contact_position} holds for any contacting
object, and with several contacts it gives their combined CoP.
Fig.~\ref{fig:handwriting}d shows the paths, drawn whenever
$|\hat{F}_z|\geq 1$~N. We will
evaluate the tracking against a motion-capture reference in future work.

\subsection{Fingertip Sensors for Pinch Grasping}

We demonstrate EddyTac in a smaller form factor on a fingertip sensor by scaling down the sensor design (Fig.~\ref{fig:fingertip}).
Six $3.8$~mm coils lie on a ring of radius $5$~mm, with a seventh coil at the center.
Its circular head is
$16.6$~mm across. The sensor is $2.4$~mm thick including the flexible printed
circuit board (FPCB), which wraps around a 3D printed finger mount. The two inductance converters and
their capacitors face inward in a recess under the side cover, and the cable
connectors (with the same pinout for different finger directions) sit opposite the sensing head.

The thumb and index sensors were calibrated separately against the ATI reference. On a separate test session for each, the RMSE was
0.06--0.22~N for forces and 0.36--0.97~mN$\cdot$m for moments on the thumb,
and 0.04--0.12~N and 0.33--0.82~mN$\cdot$m on the index.

To test whether the fingertip moments can locate a grasp, we lifted a
$56$~g printed plastic bar once at each of five positions, from right to
left (Fig.~\ref{fig:pinch}a--e). The grasp coordinate $x_{\mathrm{g}}$ runs
from the left end of the bar ($0$~mm) to its right end ($L=150$~mm), as in
Fig.~\ref{fig:pinch}c. The targets for the head center were the quarter,
half, and three-quarter points and the two points $10.3$~mm inside the ends
($2$~mm plus the $8.3$~mm head radius).

For a uniform bar held still and level, the weight $mg$ exerts a moment of
magnitude $mg\lvert x_{\mathrm{g}}-L/2\rvert$ about the grasp axis, the line
through the two contacts. We took the local $z$ axes of the two sensors as
pointing in opposite directions along this axis, so their moments combine as
the difference
$\Delta\tau_z=\tau_{z,\mathrm{index}}-\tau_{z,\mathrm{thumb}}$. To exclude the transients at
contact and release, we defined a hold as an interval in which the force
magnitude at both fingertips exceeded $2$~N for more than $1.5$~s. We trimmed
$0.5$~s from each end of every hold and took the median of $\Delta\tau_z$ over
the rest. The position estimate is
\begin{equation}
  \hat{x}_{\mathrm{g}} = \frac{L}{2} +
   \frac{\operatorname{median}(\Delta\tau_z)-\Delta\tau_{z,0}}{mg}.
  \label{eq:bar_grasp_position}
\end{equation}
Here $mg=0.549$~N. The weight exerts no moment about a grasp at the
midpoint, so the median difference in a separate grasp there,
$\Delta\tau_{z,0}=2.85$~mN$\cdot$m, is an offset between the two sensors,
which we subtract from every grasp. With moments in mN$\cdot$m and the
weight in N, \eqref{eq:bar_grasp_position} gives $\hat{x}_{\mathrm{g}}$ in mm.

The estimates differed from the five targets by $4.3$~mm on average.
The midpoint estimate was $73.4$~mm for the
$75.0$~mm target, and both outer estimates fell toward the midpoint by
$10.5$~mm on the right and $2.9$~mm on the left. Because neither the
finger orientations nor the actual head centers were tracked, the $4.3$~mm
compares the estimates with the intended placements only.

Toward either end, the grip force roughly doubled, to \mbox{$8$--$10$~N} at
the outer grasps (Fig.~\ref{fig:pinch}f). There
the weight exerts a larger moment about the grasp axis, and holding the bar
without rotational slip takes a larger normal force, since the friction
limit surface of each contact scales with the normal force~\cite{goyal1991}.
The normal forces were similar at the two outer grasps, but the local
moments of both fingers reversed sign between them
(Fig.~\ref{fig:pinch}g).

\section{Conclusion}
\label{sec:conclusion}

This paper presents EddyTac, a thin six-axis tactile sensor made of coplanar PCB
coils and a silicone stack cast in 3D-printed molds. The edges of the conductive
plate are placed and oriented so that its translations and rotations give
distinguishable coil responses, and a learned calibration maps these
responses to three force and three moment components. The sensor needs no
upright sensing elements, and its silicone stack carries no wiring. In the test session, its RMSE was
0.09--0.26~N for forces and 1.0--2.1~mN$\cdot$m for moments.

The estimated wrench also gives the CoP of any contact on
the sensor surface, and it traced the strokes of seven handwritten letters.
With fingertip sensors on the thumb and index finger, the normal
forces were similar when a bar was held near either end, but the difference
between the two moments about the grasp axis changed sign. A static moment
balance located the five grasps along the bar with a mean absolute
difference of 4.3~mm from the intended positions.

EddyTac has three main limitations. We calibrated and tested EddyTac only
under compression, and the silicone's slow creep causes some drift between sessions.
A nearby metal object can affect the coil readings. A nonconductive spacer can reduce
this effect by increasing the distance between the object and the coils.
A load must also move the conductive plate, so a press
between the blades, where no plate lies under the silicone, is not measured
reliably. Future design experiments will explore addressing these limitations. 

Future work will characterize temperature effects, hysteresis, and
durability. We will make the conductive plate from copper-clad FR4, with the
blades etched on a rigid disk that spans the sensing area, so that a press
between the blades also moves the plate. We plan to arrange the sensors in a grid to map contact loads
at finer spatial resolution, and to put motion capture markers on the
fingertip mounts so that human demonstrations record finger motion together
with contact forces and moments. We also plan to release the design files,
fabrication instructions, and calibration tools.

\end{document}